# Object Detection based on the Collection of Geometric Evidence

Wei Hui        Tang Fu-yu

*Abstract*—**Artificial objects usually have very stable shape features, which are stable, persistent properties in geometry. They can provide evidence for object recognition. Shape features are more stable and more distinguishing than appearance features, color features, grayscale features, or gradient features. The difficulty with object recognition based on shape features is that objects may differ in color, lighting, size, position, pose, and background interference, and it is not currently possible to predict all possible conditions. The variety of objects and conditions renders object recognition based on geometric features very challenging. This paper provides a method based on shape templates, which involves the selection, collection, and combination discrimination of geometric evidence of the edge segments of images, to find out the target object accurately from background, and it is able to identify the semantic attributes of each line segment of the target object. In essence, the method involves solving a global optimal combinatorial optimization problem. Although the complexity of the global optimal combinatorial optimization problem seems to be very high, there is no need to define the complex feature vector and no need for any expensive training process. It has very good generalization ability and environmental adaptability, and more solid basis for cognitive psychology than other methods. The process of collecting geometric evidence, which is simple and universal, shows considerable prospects for practical use. The experimental results prove that the method has great advantages in response to changes in the environment, invariant recognition, pinpointing the geometry of objects, search efficiency, and efficient calculation. This attempt contributes to understanding of some types of universal processing during the process of object recognition.**

## I. INTRODUCTION

IN daily life, we almost always need to recognize the objects that we see. Object recognition typically occurs so effortlessly that it is hard to believe it is actually a rather complex achievement. In the area of cognitive psychology, experts believe that there are three key processes involved in object recognition [6]:

First, there are usually numerous different overlapping objects in the visual environment, and the viewer must somehow decide where one object ends and the next starts. This is a difficult task.

Second, objects can be recognized accurately over a wide range of viewing distances and orientations. The apparent size and shape of an object do not change despite large variations in the size and shape of the retinal image.

Third, allocating diverse visual stimuli to the same category of objects. Objects, such as cups, vary enormously in their visual properties (e.g., color, size, and shape), but viewers are still able to recognize them without any apparent difficulty.

Over the years, there have been plenty of theories about how we observe and recognize objects. Irving Biederman maintained that viewers recognize some basic geometries first and then later identify the object [7]. Besides, when we observe or imagine an object, we trend toward a perspective slightly above the object looking down and offset a little to the right or left. This has been dubbed the canonical perspective. Almost no one would like a perspective just above the object. Objects are recognized most quickly at this canonical perspective. At some special perspectives, such as just above the object, we might take much longer time to recognize it, and even sometimes we are unable to recognize it (e.g., some cylindrical objects) [8][9].

It's a very natural idea to recognize objects by their geometric features, and it also conforms to human cognitive behavior. In daily life, many objects (e.g., cups, basins, and buckets) differ in color and texture, and the key means of identifying them tends to involve their geometric features. Geometric features are also the most intuitive and easiest to understand. And it's also simple to implement and highly efficient. According to Biederman's theory, object recognition depends on edge information rather than on surface information (e.g., color). To test this, participants were presented with line drawings or full-color photographs of common objects for between 50 and 100 ms. Performance was comparable with the two types of stimuli: mean identification times were 11 ms faster with the colored objects, but the error rate was slightly higher. Even objects for which color would seem to be important (e.g., bananas) showed no benefit from being presented in color [6][7]. Sanocki et al. (1998) also pointed out that edge-extraction processes are more likely to lead to accurate object recognition when objects are presented on their own rather than in the context of other objects [30].

However, there are still some difficulties in object recognition based on geometric features. These are described in three key processes involved in object recognition in cognitive

This work was supported by NSFC project (Project No.61375122).

Wei Hui is with the Laboratory of Cognitive Algorithms and Modeling, Shanghai Key Laboratory of Data Science, Department of Computer Science, Fudan University, Shanghai 201203, China (e-mail: weihui@fudan.edu.cn ).

Tang Fu-yu are with Laboratory of Cognitive Algorithms and Modeling, Shanghai Key Laboratory of Data Science, Department of Computer Science, Fudan University, Shanghai 201203, China.

psychology. First, objects might be located in different scenes, and it's hard to decide where one object ends and the next starts. Second, an object could be observed in different perspectives and distances, which causes various of changes in geometric features. Third, although objects in the same category may have similar shapes, those shapes may still vary. In addition, regarding human behavior, it might be hard or even impossible to recognize objects in some special perspectives.

To be as consistent as possible with the results of psychological experiments and to help solve those difficulties, template methods that treat the geometric features in some major perspectives as templates were here used to recognize objects with corresponding templates. Here are some advantages of template methods:

First, geometric templates contribute to overcoming interference from various changes in lighting, color, and texture.

Second, geometric templates contribute to the implementation of invariant recognition at transformations such as translation, rotation, and scale, and strengthen the adaptability of each template.

Third, geometric templates contain geometric and topologic features. These features are stable.

Fourth, we just need a few major templates to recognize an object, and these do not require ex-period training, which can be costly.

Fifth, the recognition processes using geometric templates are closely connected to other cognitive processes, such as inductive learning, knowledge representation, reasoning, attention selection, and hypothesis testing.

Sixth, recognition using geometric templates is simple to implement, and it is also efficient.
In section 2, some typical template methods for object recognition are introduced. In section 3, the template is defined and preprocessing is performed. In section 4, the evidence used for object recognition is defined and a search for such evidence searching is conducted. In section 5, template matching is performed and then the results are filtered. Finally, the experimental results and conclusion are shown in sections 6 and 7.

## II. Overview

Object Recognition based on template methods can be categorized as follows:

The first category contains methods based on contours or shapes, which have been described in many previous works [2][3][17][18][19][20][21][25]. Methods in this category generally obtain templates based on contours or shapes from sample images or other ways, and then compare these templates with the contours or shapes in target image. Changes in lighting and color do not much affect the methods. However, they are sensitive to interference lines in the background. If there are many interference lines in the background, the recognition rate and efficiency decrease greatly.

The second category includes methods based on color or grayscale [1][11][12][26]. These methods work at pixel level, consider both the position and strength of pixels, and then find the difference between templates and target images. These methods perform well on objects that have significant color and grayscale features, such as faces. However, they are sensitive to changes in lighting and color. Therefore, they do not perform well for objects that change in lighting and color.

The third category includes methods based on texture and gradient [4][22][27][28][29]. These methods, which consider the textural features of objects, compare the textures of templates and target images for object recognition. They perform well on objects that have significant textural features. However, they are also sensitive to changes in lighting and color.

There are also other methods, such as those based on histograms of receptive field responses [13][14][15][16]. However, all methods need templates of some sort to perform object recognition. An object can be observed from any perspectives or at any distances, so may be necessary to collect multiple templates for the same object. The retinal image of an object may differ, so new shapes can be produced through any type of transformation, such as translation, rotation, or changes in scale. In general, traditional template methods only recognize objects that have shapes similar to those of their templates. To perform object recognition for various of shapes, new templates should be constructed for any new shapes produced through transformation, but this is obviously quite undesirable. And the reason is that these methods do not make full use of the geometric features of objects and templates. The templates of objects can also be used to perform two-dimensional transformations, such as translation, rotation, and changes in scale. However, these methods usually consider only transformations of translation and scale, or even fail to consider this. In this way, these methods require many templates because of the low adaptability of templates. Based on this issue, this paper describes a new method that makes fuller use of the geometric features of objects and templates to improve the flexibility and adaptability of each template. This method acts on artificial objects that are composed primarily of line segments. These objects are very common, and they can be seen anywhere in our daily life.

## III. Template definition and data pre-processing

The method described here is based on geometric features of objects, which include points, lines, angles, position, topological location, etc. All of these features can be represented and calculated using points and lines. Here, a set of line segments serves as a template for the target object. Fig. 1 shows a template of F117 tilted and at a slight angle above. This is based on a set of line segments. This type of template is not only simple and intuitive but also reflects the geometric features of objects very well.

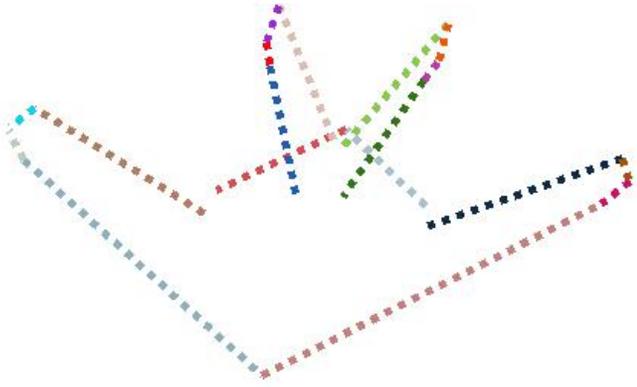

Fig. 1. A template of F117 based on a set of line segments.

Similarly, the preprocessing for the original images is very simple, and it includes the following steps:

The first step, perform edge detection for the original image, such as Canny edge detection or Berkeley edge detection [23].

The second step, get line segments for the result of edge detection, such as Hough transform for line detection or edge link method [24].

The third step, combine the line segments, which are continuous at position, to a longer line segment.

Fig. 2 shows the results based on Berkeley edge detection and edge link method.

## IV. EVIDENCE SEARCHING

### A. Hypothesis generation

After getting the line segments of the templates (Fig. 1) and target images (Fig. 2), the problem of object recognition becomes the problem of combining line segments. This shifts the problem of object recognition from the pixel level to the geometric line segment level. At this level, all geometric features of objects are well represented, and they can be calculated very simply and efficiently. In general, an image includes thousands of pixels. At the pixel level, it is not efficient to perform recognition, and it is also difficult to extract useful information, such as geometric features and topologic features. However, at the line segment level, a target image usually includes merely hundreds of line segments, and a template can include fewer than 100 line segments. It is not only easier to find and calculate geometric features but also possible to perform more complex calculations, such as geometric transformation and hypothesis testing. We need only select a series of line segments based on geometric features of

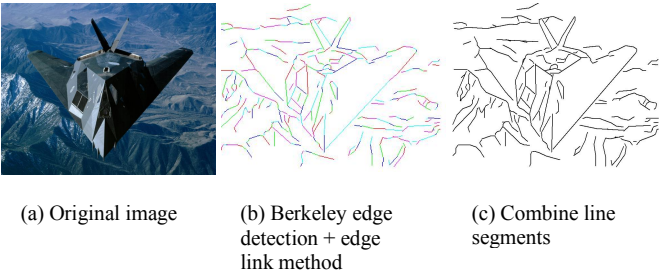

(a) Original image  (b) Berkeley edge detection + edge link method  (c) Combine line segments

Fig. 2. Preprocessing of the original image.

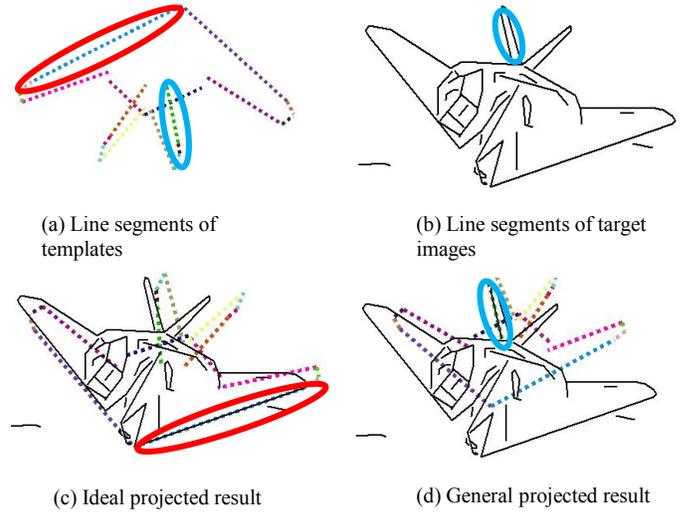

(a) Line segments of templates   (b) Line segments of target images

(c) Ideal projected result   (d) General projected result

Fig. 3. Projection of the line segments from the template onto the target image.

templates, which are derived from target images, and then combine them so that they correspond with the line segments of the templates.

For this purpose, the line segments of templates should align with the line segments of target images. That means we need to perform transformations, such as translation, rotation, and changes in scale for the line segments of templates, to produce template line segments that correspond to the target images, and then project them onto the line segments of target images to produce hypotheses regarding the position, size, and shape of the objects. As shown in Fig. 3, a line segment was selected randomly from a template, here marked with red or blue ellipse (Fig. 3(a)). A line segment was also selected randomly from the target image, and marked with red or blue ellipse (Fig. 3(b)). We assume that the two line segments marked with ellipses of matching color may correspond to each other, and then the line segments of the template can be transformed and projected onto the target image (Fig. 3(c) and 3(d)).

If the transformation factors are known early on, we can perform translation, rotation, and changes in scale on the line segments of templates easily, and then project them onto the target images. However, the transformation factors are often unknown, and it is a difficult task to find suitable transformation factors. Here, we introduce a simple and efficient algorithm based on hypothesis generation:

The first step, $\forall$ $m1 \in$ the line segments of the template, such as the line segment marked in red/blue ellipse, as shown in Fig. 3(a).

The second step, $\forall$ $l1 \in$ the line segments of the target image, such as the line segment marked in red/blue ellipse, as shown in Fig. 3(b).

The third step, let $t1$ denote the transformation factor, and define *transform(m1, t1)*, which is a function of two variables about $m1$ and $t1$, as the result of transform $m1$ by $t1$. Assume *transform(m1, t1)* = $l1$, and this produces $t1$.

The fourth step, according to $t1$, transform the line segments of the template and then project them onto the target image, as

$$D_1 = |\sin \theta|;$$

$$D_2 = \frac{d}{|AB|};$$

$$D_3 = \begin{cases} 0.5, & \text{if } t_E \approx t_F \approx 0 \text{ or } t_E \approx t_F \approx 1 \\ 0, & \text{if } 0 < t_E < 1, 0 < t_F < 1 \\ 0, & \text{if } t_E \leq 0, t_F \geq 1 \text{ or } t_E \geq 1, t_F \geq 0 \\ |t_E - t_F| + \min(t_E, t_F) - 1, & \text{if } t_E, t_F > 1 \\ |t_E - t_F| + (0 - \max(t_E, t_F)), & \text{if } t_E, t_F < 0 \\ 1 - |\max(t_E, t_F)|, & \text{if } t_E < 0 < t_F < 1 \text{ or } t_F < 0 < t_E < 1 \\ |\min(t_E, t_F)|, & \text{if } 0 < t_E < 1 < t_F \text{ or } 0 < t_F < 1 < t_E \end{cases};$$

$$dis_{CD \to AB} = D_1 + D_2 + D_3;$$

if $dis_{CD \to AB} \leq TH$ and $\left(D_1, D_2, D_3 \leq \frac{2}{3} TH\right)$ then return TRUE.

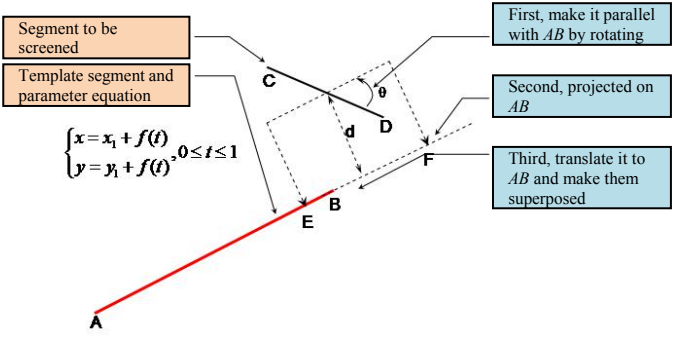

Fig. 4. Determine whether *CD* is an evidence line segment of *AB*.

(1)

shown in Fig. 3(c) and 3(d).

Here we use the geometric relationship between the line segments of templates and target images to enumerate transformation factors. Therefore, we have got many projected results, and Fig. 3(c) and 3(d) show two such results. What do we care is that, is there an ideal projected result among so many projected results? Thanks to the enumerated method of transformation factors, if there exist an ideal projected result between the template and the target image and an almost complete line segment in the target image, then this ideal projected result (or a very similar one) is among our projected results. Therefore, our method can almost always produce ideal transformation factors. Based on the experimental results, it often costs about 1 second to enumerate all possible hypotheses and then verify them. If there are too many line segments in background, it may take a few seconds.

### B. Hypothesis verification

After getting so many projected results, we need to verify these results, to find out the ideal projected result(s). During verification, the key step is to decide how to select line segments from the target image, to be corresponding with the template. The projected template is here considered a hypothesis, and we need to find some evidence to support this hypothesis. We call this process evidence collection, and call those selected line segments evidence line segments, as the evidence to support the hypothesis. During this process, we need to select line segments from the target image, and determine which line segments could be considered as the evidence line segments of a template line segment. We try to consider the problem from angle, distance, and projected position between line segments of the template and the target image, and try to make full use of geometric features. Fig. 4 shows the process. Here *CD* is a line segment from the target image, and *AB* is a template line segment.

Combined with the process in Fig. 4, here is the formula to determine whether *CD* is an evidence line segment of *AB*:

In this formula, the closer $dis_{CD \to AB}$ is to 0, the more similar the two line segments are, and then the more likely that *CD* is an evidence line segment of *AB*. *TH* is the threshold of evidence selection, commonly within the range of 0.4 to 0.6. The formula considers the relationships of angle (*D1*), distance (*D2*), and projected position (*D3*) between *CD* and *AB* at the same time.

Fig. 5 shows some examples of evidence selection, where the red dashed line is the projected template line segment (here called a hypothetical line segment), and the red solid lines are the evidence line segments of the hypothetical line segment.

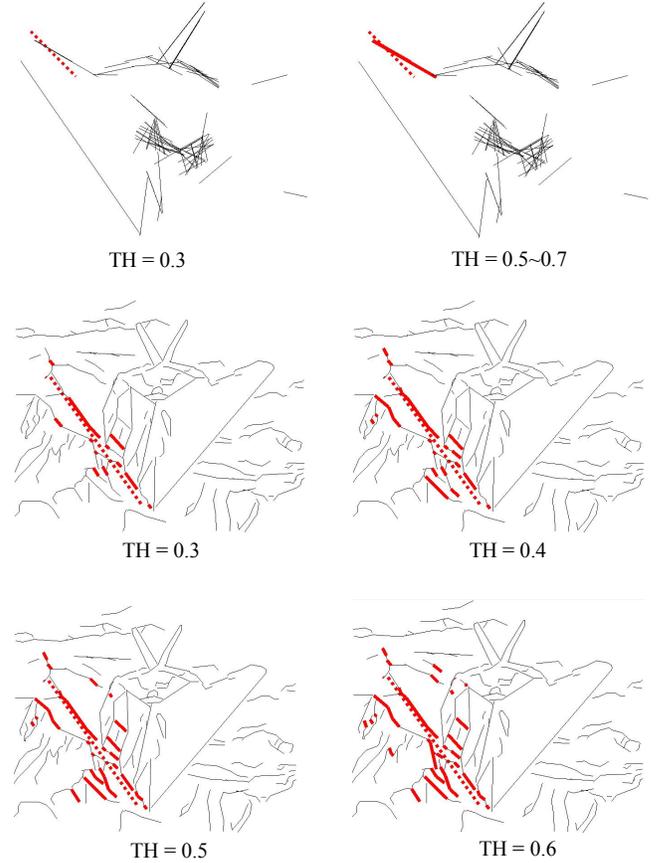

Fig. 5. Examples of evidence selection.

## V. TEMPLATE MATCHING

### A. Calculation of similarity

For each projected result, a set of evidence line segments are found for the hypothetical line segments. Here, we denote the hypothetical line segment as $h1$, and the set of evidence line segments for $h1$ as $Ev(h1)$, to serve as evidence to support this hypothetical line segment. We notice that a line segment of the target image can serve as the evidence for at most one hypothetical line segment. However, a hypothetical line segment can have one or more evidence line segments or none at all.

The closer $dis_{CD \to AB}$ is to 0, the more similar the two line segments are. In this way, if a line segment of the target image can become evidence for more than one hypothetical line segments, this line segment should serve as evidence for the most similar hypothetical line segment, for which $dis_{CD \to AB}$ has the lowest value. We denote the set of hypothetical line segments as $\{h_1, h_2, ..., h_M\}$, the set of line segments of the target image as $\{l_1, l_2, ..., l_N\}$, and the threshold of evidence selection as $TH$. The pseudo-code of the algorithm used to find the set of evidence line segments is as follows:

```
Let Ev(h₁) = Ev(h₂) = ...= Ev(hₘ) = Ø
for each lᵢ in {l₁, l₂, ..., lₙ} do begin
    h_best ← NULL; mindis ← TH;
    for each hⱼ in {h₁, h₂, ..., hₘ} do begin
        Calculate the dis_{CD→AB} about lᵢ and hⱼ
        if (( dis_{CD→AB} satisfies the discriminant) and
    ( dis_{CD→AB} < mindis)) then begin
            h_best ← hⱼ;
            mindis ← dis_{CD→AB} ;
        end if
    end for
    if (h_best != NULL) then begin
        Let lᵢ join in Ev(h_best)
    end if
end for
```

After producing the set of evidence line segments, we regard them as an overall shape, and then calculate the similarity between the set of evidence line segments and the set of hypothetical line segments. The formula used to calculate similarity shows as follows:

$$matchval(h_j) = \min\left( len(h_j), \sum_{l \in Ev(h_j)} matchlen(l, h_j) \right), j = 1, 2, \cdots M$$

$$Sim = \frac{\sum_{j=1}^{M} matchval(h_j)}{\sum_{j=1}^{M} len(h_j)} \cdot \frac{\gamma}{M} \tag{2}$$

Here, we define as the length of $h_j$, $matchlen(l, h_j)$ as the matched length between $h_j$ and the transformed $l$ (as shown in Fig. 4), and $\gamma$ as the number of hypothetical line segments that are supported by evidence. The value of $Sim$ belongs to [0, 1], and the bigger the value, the more similar the set of evidence line segments and hypothetical line segments.

### B. Filtering of results

After calculating the value of $Sim$ for each projected result, all projected results can be sorted in descending order of similarity. To obtain more accurate results, we need to perform further filtering for those projected results with high ranking (such as the top 10 projected results). Then those bad projected results can be eliminated. Such results have high similarity values but are composed of background line segments and other interference line segments.

Here, filtering can be performed on those high-ranking projected results by the area size, which is a geometric feature of the object.

If a projected result is good, then the size of its area composed of the set of evidence line segments should be more than half of the set of hypothetical line segments. Therefore, we can calculate and then compare the area size of them. The process is as follows:

The first step, calculate the center point of the set of hypothetical line segments. This is defined as $CenterP = (CenterP_x, CenterP_y)$, where

$$CenterP_x = \frac{\sum_{j=1}^{M} h_{xj}}{M}$$

$$CenterP_y = \frac{\sum_{j=1}^{M} h_{yj}}{M} \tag{3}$$

The second step, calculate the area size of the set of hypothetical line segments:

$$Area_{Hy} = \sum_{j=1}^{M} Triangle(h_j, CenterP) \tag{4}$$

where $Triangle(h_j, CenterP)$ denotes the area size of triangle composed by $h_j$ and $CenterP$.

The third step, calculate the area size of the set of evidence line segments:

$$Area_{Ev} = \sum_{j=1}^{M} func(j) \tag{5}$$

where

$$func(j) = \begin{cases} 0 & Ev(h_j) = \varnothing \\ Triangle(h_j, CenterP) & Ev(h_j) \neq \varnothing \end{cases} \tag{6}$$

The fourth step, compare the sizes of the two areas. If $Area_{Ev} \geq \frac{1}{2} Area_{Hy}$, then it is a good projected result. Otherwise, it is a bad projected result and should be eliminated.

## VI. EXPERIMENTAL RESULTS

### A. Our experimental results

First, our program was run on some real images. Fig. 6 lists some results. In its column of the detected object, those highly suspected lines are colored, and the template that best matches the combination of colored lines is drawn with dotted lines. The hypothesis line and its evidence line are drawn with the same color. From these areas, we can see that F117s are detected correctly, and each colored line is assigned the correct structural role in the template.

Then, we compared our method with several state-of-the-art methods claimed to be shape-based.

| Original image | Line image | Detected object | Original image | Line image | Detected object |

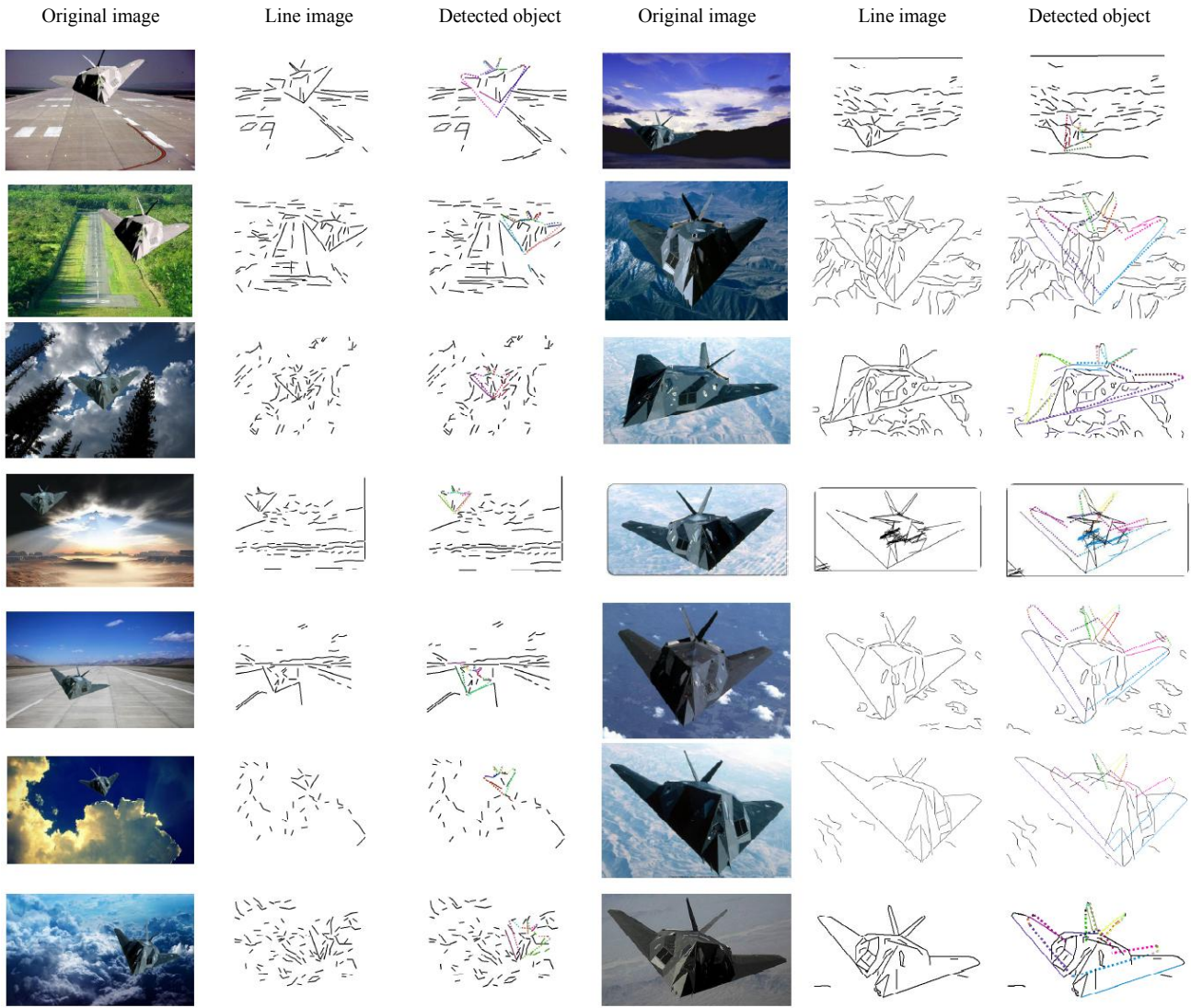

Fig. 6. Detection results of F117s.

### B. Comparison with other methods

#### 1) Fan-shape model

In CVPR2012, a Fan-shape model for object detection [31] was proposed, and we compared with it. First, besides that

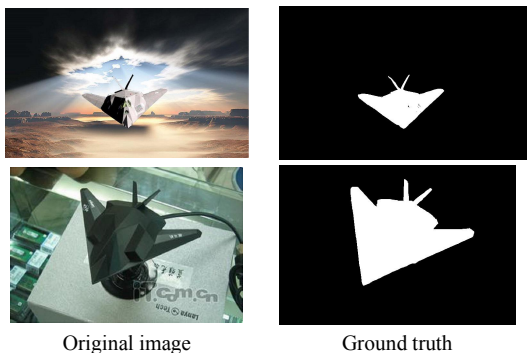

Original image      Ground truth

Fig. 7. Two pairs of image samples in the newly created dataset, F117 dataset.

famous ETHZ dataset, we designed a new dataset, F117 dataset, in which there are 50 images for training (our method does not need training, and this is for Fan-shape model) and another 121 images for testing. Fig. 7 shows two samples in this dataset. The Fan-shape model was trained on this new dataset.

After training, we ever expected that this shape-based model can work well on those clean, standard images, such as images with the perfect contour only (in the right of Fig. 8). Pure contour information is always thought to be enough for shape recognition, but the actual fact is that the Fan-model failed completely, i.e. it found nothing in these two almost perfect testing images, this is because this model actually depends more on SIFT features, however line images can not provide gray or color gradient information. Another disproof is listed in the third row of Fig. 8, in which we shown a giraffe image upon which the Fan-model functioned normally. Then we replaced this original image by its grayscale and color-inversing versions, as a consequence the Fan-shape model method failed. These results once again doubled the generality of this method.

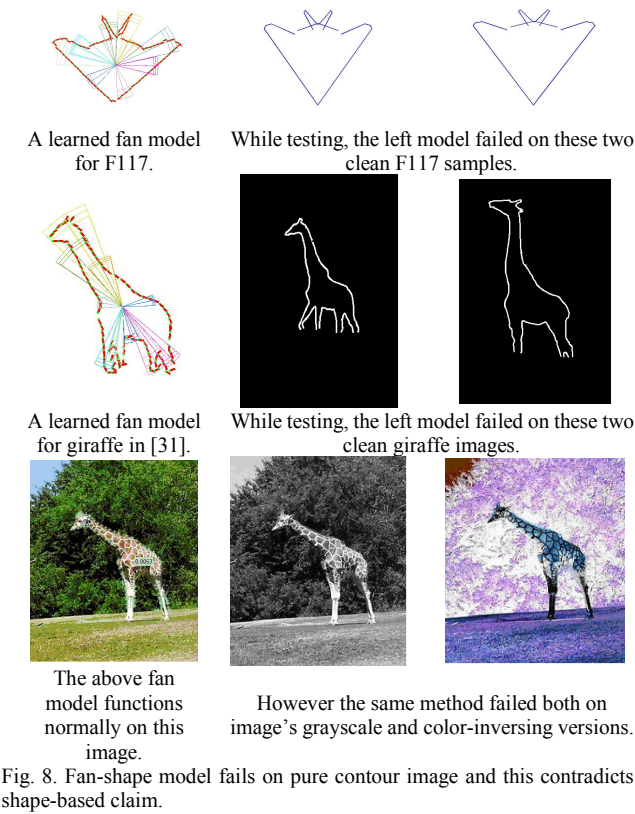

A learned fan model for F117.

While testing, the left model failed on these two clean F117 samples.

A learned fan model for giraffe in [31].

While testing, the left model failed on these two clean giraffe images.

The above fan model functions normally on this image.

However the same method failed both on image's grayscale and color-inversing versions.

Fig. 8. Fan-shape model fails on pure contour image and this contradicts its shape-based claim.

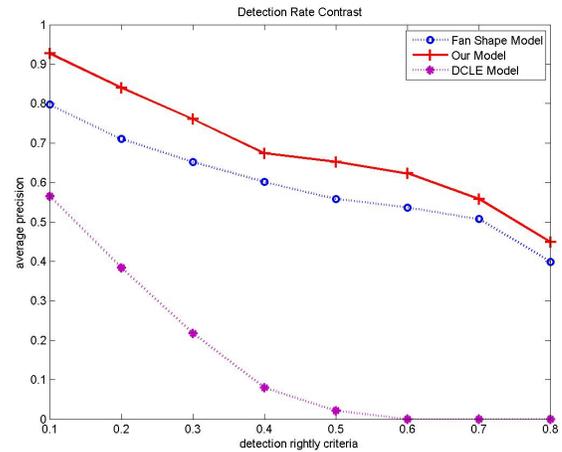

(a) Detection rate comparison

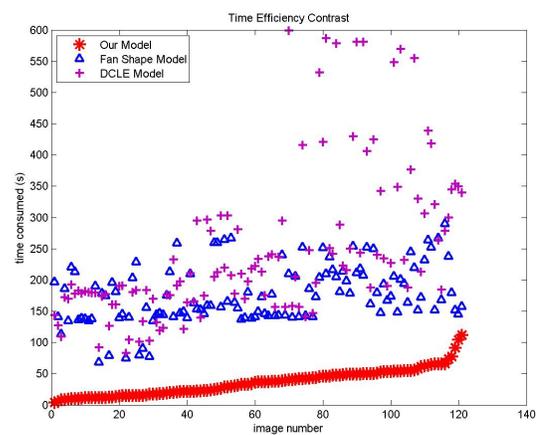

(b) Time-costing comparison

Fig. 9. Performance comparisons among our method, the Fan-shape model, and DCLE module.

After training, we ever expected that this shape-based model can work well on those clean, standard images, such as images with the perfect contour only (in the right of Fig. 8). Pure contour information is always thought to be enough for shape recognition, but the actual fact is that the Fan-model failed completely, i.e. it found nothing in these two almost perfect testing images, this is because this model actually depends more on SIFT features, however line images can not provide gray or color gradient information. Another disproof is listed in the third row of Fig. 8, in which we shown a giraffe image upon which the Fan-model functioned normally. Then we replaced this original image by its grayscale and color-inversing versions, as a consequence the Fan-shape model method failed. These results once again doubled the generality of this method.

For the performances of different methods that work on classical ETHZ dataset are mature, we altered to the new F117 dataset and collected statistical datum. Fig. 9 shows the detection rate comparison and time-costing comparison among our method, the Fan-shape model [31], and the DCLE module [32][33]. In Fig. 9, detection rightly criteria means a detection is deemed correct if the intersection of the detected bounding box and ground truth over the union of the two bounding boxes is larger than the x axis value. From these curves, we can see that our method not only performs better but also spends much less time.

## 2) Discriminative combinations of line segments and ellipses (DCLE)

A.S. Chia et al. described a method based on discriminative combinations in CVPR2010 [32] and PAMI2012 [33]. This

Detection result on original images

Detection result on re-organized images

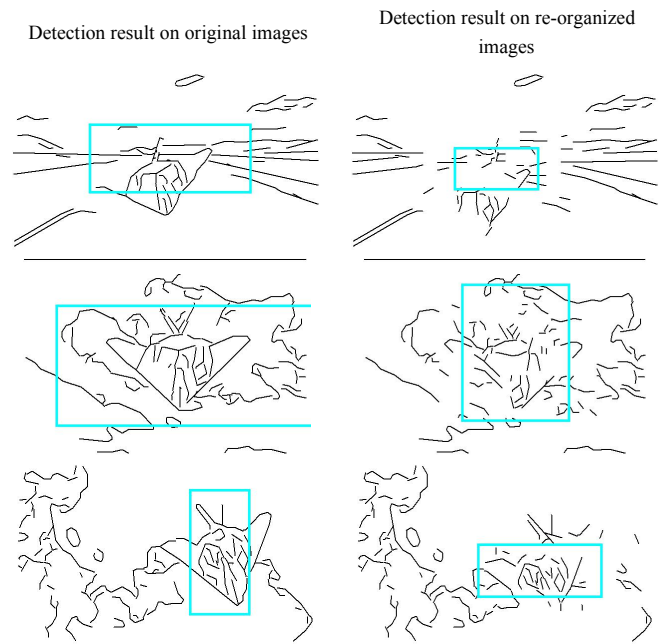

Fig. 10. Results of the method based on discriminative combinations.

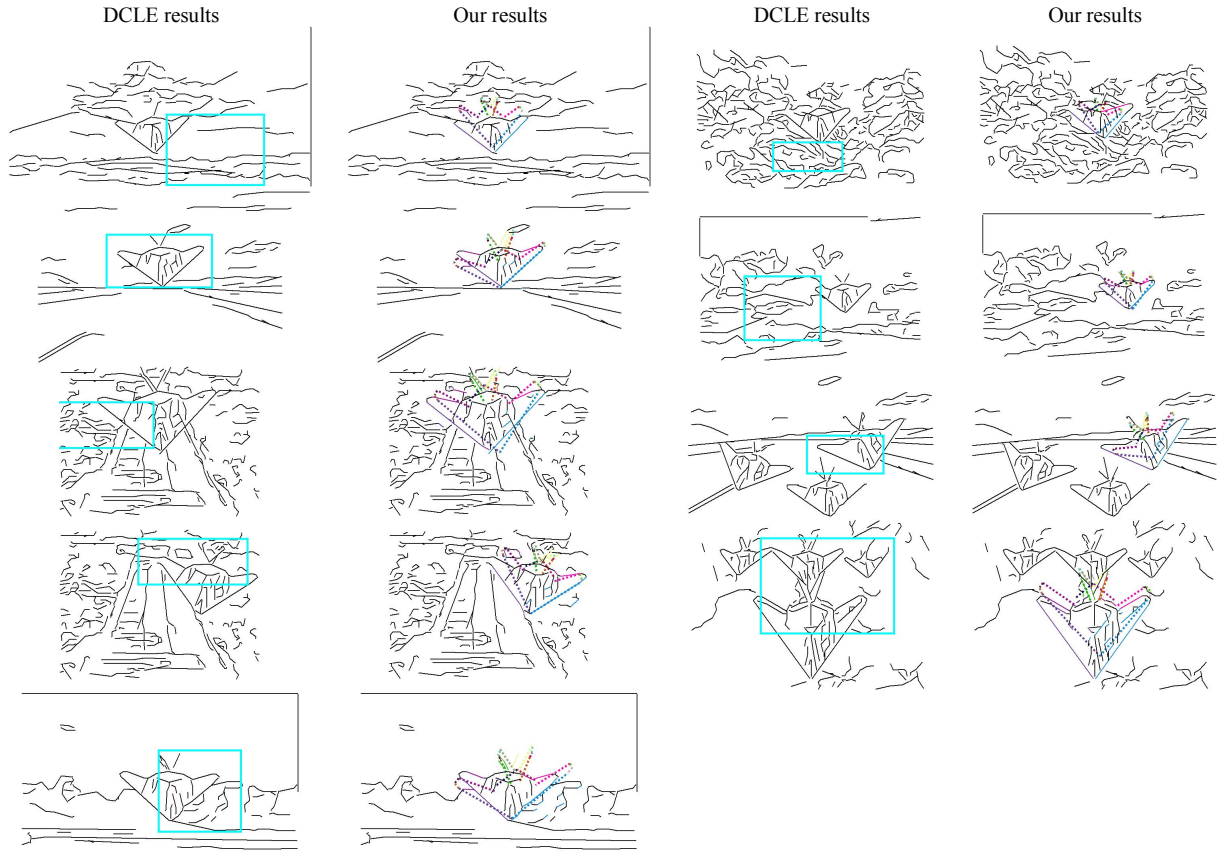

Fig. 11. Comparison between our method and the method based on discriminative combinations.

method is characterized by the use of line segment primitives, ellipse primitives, and SIFT features. The method pairs a reference primitive to its neighboring primitive to construct more discriminative shape-tokens, and then get the codebook of the target object (adding SIFT features if necessary). After that, it learns from the codebook and produces a series of discriminative codeword combinations for object recognition. In Fig. 10, the left panel shows some results for images of F117 based on this method, and the right panel shows results for re-organized images, which cut objects into small pieces and then combine them randomly. In theory, the method should not be able to detect out the F117 on re-organized images. However, in fact, the method still detected out the F117 in its original position.

Our method is able to work well on pure shape images, so there is no need for any SIFT features. In order to compare the two methods fairly, we experiment on images of the edge segments. The experimental results are shown in Fig. 11.

### C. Results on UIUC car datasets

In order to compare with more methods, we test our method on UIUC car datasets. Images of the datasets are quite small with the fuzzy object boundaries. Besides, images are grey and difficult to extract shape boundaries. Therefore, this is a challenge to our algorithm. We used a side view template to test 170 single-scale images of the UIUC datasets. Fig. 12 shows some experimental results, and we can see that our algorithm detected the cars very accurately.

To test detection accuracy, for each test image, we regard those results with high ranking and similarity above a certain threshold as detection results. If the intersect area between a detection result and a ground truth was more than a half, the result was considered to have found the car at corresponding position of that ground truth correctly. We calculated two precision-recall values, here one had maximum recall value and the other had maximum precision value. Fig. 13 shows the precision-recall results of our algorithm and some other methods.

As shown in Fig. 13, we calculated the recall-precision equal error rate (EER) for our algorithm on UIUC car datasets. Table 1 shows the EER of our algorithm and some advanced methods, and we can see that our algorithm based on geometric features are as thoroughly able to produce ideal results as other methods.

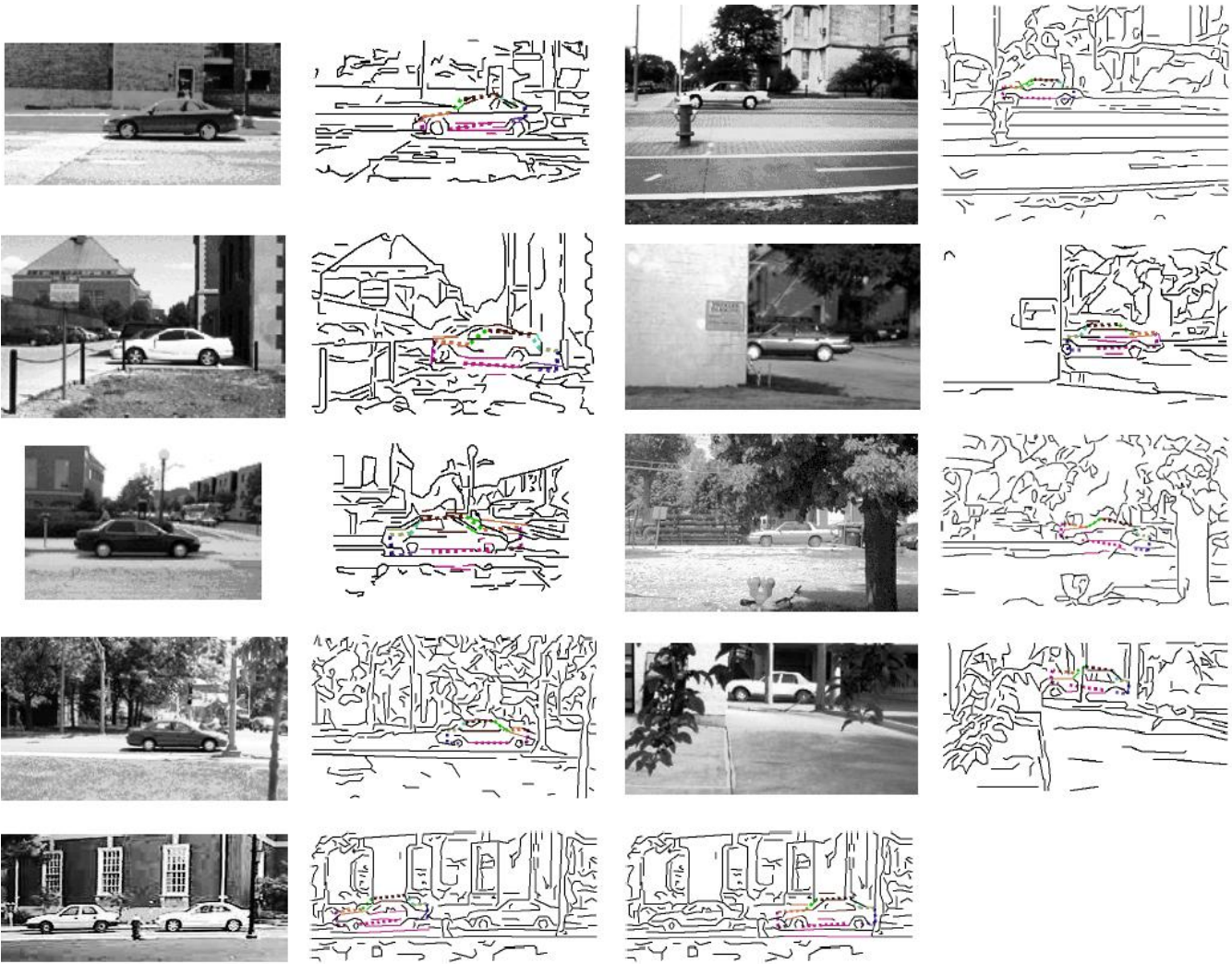

Fig. 12. Our results on UIUC car datasets.

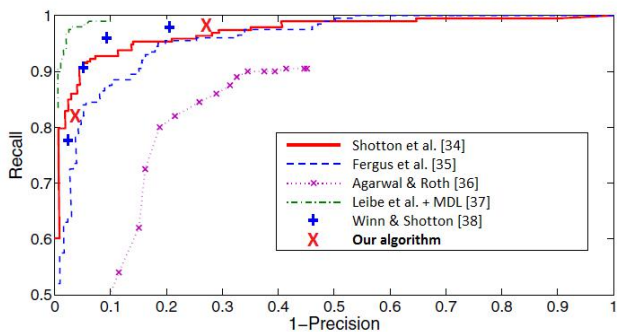

Fig. 13. Recall-precision curves for the UIUC datasets.

TABLE I
PERFORMANCE OF DIFFERENT METHODS ON THE UIUC CAR DATA SETS AT
RECALL-PRECISION EQUAL ERROR RATE (EER)

| Methods | UIUC-Single |
|---|---|
| Hough-based methods | |
| Implicit Shape Model | 91% |
| ISM+verification | 97.5% |
| Boundary Shape Model | 85% |
| Max-margin HT+verif. | 97.5% |
| Random forest based method | |
| LayoutCRF | 93% |
| State-of-the-art | |
| Mutch and Lowe CVPR2006 | 99.9% |
| Lampert et al. CVPR2008 | 98.5% |
| Karlinsky et al. CVPR2010 | 99.5% |
| Hough Forests | |
| Hough Forest | 98.5% |
| HF-Weaker supervision | 94.4% |
| HF-Sparse | 95.5% |
| HF-Weighted | 98.5% |
| Geometric based method | |
| Our algorithm | 96% |

## VII. Conclusions

At present, the most common method of object recognition is the pattern classification method based on machine learning. This method has the following problems. First, the feature descriptors of images have poor generality, and the feature vectors often have very high dimensions, which creates huge computational load. Second, the machine learning algorithms are complex, with strong application specificity and data-set specificity, and the classifiers are also hard to generalize. Third, the classification criteria produced by machine learning algorithms are obscure and have poor declarative semantics. These criteria are often implicit and hard to present as knowledge. Finally, it reduces the recognition problem to a classification problem and conceals many inner processes. Based on the above viewpoint, this paper hopes to use a simpler and more natural method of recognition. We believe that search and optimization have such attributes.

An outstanding advantage of our method is the simplicity of the algorithm. In the principle, we use the template method. In the image representation, we only use line segments, and there is no need for other complex and high-dimensional feature descriptors. In the recognition process, we collect the contour line segments that satisfy the geometrical constraint as evidence, and there is no need for the expensive machine learning processes to train and produce classifier. In the numerical calculation, it only involves some simple geometry and algebra calculations, which are small computation. Based on the simplicity of these components, our method still performance very well.

Purely shape-based methods provide more generalization. This paper uses only the geometric features of objects to perform object recognition, because the geometric features are the most stable features of the object and they tolerate changes to the environment, color, lighting, scale, and rotation well. In view of the fact that changes in the environment are unlimited and unpredicted, those methods, which are based on the representation of physical feature vectors and statistical learning, are difficult to produce comprehensive statistics. However, methods based on the representation of shapes and contours can contain all possible poses using only a few templates, and there is no need for expensive learning and training processes. The most important is that representation based on shapes can describe the structural features of objects well, and the accuracy of semantic representation of objects is better than the classifiers based on positive and negative rules.

Evidence accumulation has simpler and more general implementation than other methods. Object recognition processes based on geometric evidence are more similar to the human cognitive process, which mainly includes evidence collection, hypothesis formation, and hypothesis verification. This process is simpler and more general, and it includes the most basic steps of recognition. Although this paper does not consider color features, texture features, or depth features, these can be added to improve the efficiency of the combination process by serving as clues for how to combine line segments.

## References

[1] T. Gevers, A. Smeulders. "Color Based Object Recognition". Pattern Recognition, 1997

[2] Alexander C. Berg, Tamara L. Berg, Jitendra Malik. "Shape matching and object recognition using low distortion correspondence". In CVPR, 2005

[3] Serge Belongie, Jitendra Malik, Jan Puzicha. "Shape Matching and Object Recognition Using Shape Contexts". IEEE Transactions on Pattern Analysis and Machine Intelligence, 2001

[4] Thomas Serre, Lior Wolf, Stanley Bileschi, Maximilian Riesenhuber, Tomaso Poggio. "Robust object recognition with cortex-like mechanisms". IEEE Transactions on Pattern Analysis and Machine Intelligence, 2007

[5] David R. Martin, Charless C. Fowlkes, Jitendra Malik. "Learning to detect natural image boundaries using local brightness, color, and texture cues". PAMI, 2004

[6] Eysenck, M. W. and Keane, M. T. "Cognitive Psychology: A Student's Handbook. 4th edn". Hove, UK: Psychology Press, 2000

[7] Biederman I. "Recognition-by-components: A theory of human image understanding". Psycho Rev, 1987, 94: 115–147

[8] Susan M. Weinschenk. "100 Things Every Designer Needs to Know About People". Pearson Education, 2011

[9] Stephen R. Palmer and Paul Chase. "Canonical Perspective and the Perception of Objects". LEA, 1981, p. 135-151.

[10] Hubel, D. H. and T. N. Wiesel. "Receptive Fields Of Single Neurones In The Cat's Striate Cortex". Journal of Physiology, 1959, 148, 574-591.

[11] T. Gevers and H. Stokman, "Robust Histogram Construction from Color Invariants for Object Recognition". IEEE Trans. Pattern Analysis and Machine Intelligence, vol. 26, no. 1, pp. 113-118, Jan. 2004.

[12] Tsin, Y., Collins, R. T., Ramesh, V., & Kanade, T. "Bayesian color constancy for outdoor object recognition". In IEEE conference on computer vision and pattern recognition, 2001

[13] M. J. Swain and D. H. Ballard. "Colour indexing". International Journal of Computer Vision, 7:1, 11-32, 1991.

[14] B. Schiele and J. L. Crowley. "Recognition without correspondence using multidimensional receptive field histograms". International Journal of Computer Vision, 36:1, 31-50, 2000

[15] O. Linde and T. Lindeberg. "Object recognition using composed receptive field histograms of higher dimensionality". Proc. International Conference on Pattern Recognition (ICPR'04), Cambridge, U.K. II:1-6, 2004.

[16] O. Linde and T. Lindeberg. "Composed complex-cue histograms: An investigation of the information content in receptive field based image descriptors for object recognition". Computer Vision and Image Understanding, 116:4, 538-560, 2012.

[17] J. Zhang , X. Zhang , H. Krim and G. G. Walter. "Object representation and recognition in shape spaces". Pattern Recogn., vol. 36, pp.1143 -1154, 2003

[18] N. Ansari and E. J. Delp. "Partial shape recognition: A landmark-based approach". IEEE Trans. Pattern Anal. Machine Intell., vol. 12, no. 5, pp.470 -483, 1990

[19] Marszalek M. and Schmid C." Accurate object localization with shape masks". In CVPR, 2007

[20] Marszalek M. and Schmid C." Accurate Object Recognition with Shape Masks". In CVPR, April 2012, Volume 97, Issue 2, pp 191-209

[21] A Zisserman, C. Schmid, K. Mikolajczyk. "Shape recognition with edge-based features". Proceedings of the British Machine Vision Conference, 2003, Vol. 2, pp. 779-788.

[22] Alin Drimus, Mikkel Børlum Petersen, Arne Bilberg. " Object texture recognition by dynamic tactile sensing using active exploration". RO-MAN, 2012 IEEE.

[23] Arbelaez P. and Maire M. and Fowlkes C.C. and Malik J. "From contours to regions: An empirical evaluation". In CVPR, pp. 2294–2301, 2009.

[24] Peter Kovesi. "MATLAB and Octave Functions for Computer Vision and Image Processing". http://www.csse.uwa.edu.au/~pk/research/matlabfns/#edgelink

[25] A. Diplaros, Th. Gevers, and I. Patras. "Combining colour and shape information for illuminationviewpoint invariant object recognition". IEEE Trans. on Image Processing, vol. 15(1), pp. 1-11, 2006.

[26] D. Muselet and L. Macaire. "Combining color and spatial information for object recognition across illumination changes". Pattern Recognition Letters, vol. 28 (10), pp. 1176-1185, 2007.

[27] Drimbarean A. and Whelan P. "Experiments in colour texture analysis". Pattern Recognition Letters, vol. 22, 10, 2001, p.1161-1167

[28] S. Lazebnik, C. Schmid and J. Ponce. "A Maximum Entropy Framework for Part-Based Texture and Object Recognition". Proc. Int',l Conf. Computer Vision, 2005.

[29] N. Dalal and B. Triggs. "Histograms of Oriented Gradients for Human Detection". Proc. IEEE Conf. Computer Vision and Pattern Recognition, vol. 2, pp. 886-893, June 2005

[30] Sanocki, T., Bowyer, K.W., Heath, M.D., & Sarkar, S. "Are edges sufficient for object recognition?". Journal of Experimental Psychology: Human Perception & Performance, 24, 1998, 340– 349.

[31] Xinggang Wang, Xiang Bai, Tianyang Ma, Wenyu Liu, Longin Jan Latecki, Fan shape model for object detection, CVPR2012.

[32] Alex Yong-sang Chia, Suaanto Rahardja, Deepu Rajan, Maylor Karhang Leung, Object recognition by discriminative combination of line segments and ellipses, CVPR 2010, 2225-2232

[33] Alex Yong-sang Chia, Deepu Rajan, Maylor Karhang Leung, Suaanto Rahardja, Object recognition by discriminative combination of line segments, ellipses, and appearance features, IEEE Transactions on Pattern Analysis and Machine Intelligence, Vol.34(9), 2012, 1758-1772

[34] J. Shotton, A. Blake, and R. Cipolla. Contour-based learning for object detection. In ICCV, 2005.

[35] R. Fergus, P. Perona, and A. Zisserman. Object class recognition by unsupervised scale-invariant learning. In Computer Vision and Pattern Recognition, 2003.

[36] S. Agarwal and D. Roth. Learning a sparse representation for object detection. In European Conference on Computer Vision, 2002.

[37] B. Leibe, A. Leonardis, and B. Schiele. Combined object categorization and segmentation with an implicit shape model. In ECCV'04 Workshop om Statistical Learning in Computer Vision, May 2004.

[38] J. Winn and J.Shotton. The Layout Consistent Random Field for Recognizing and Segmenting Partially Occluded Objects. In CVPR, 2006.

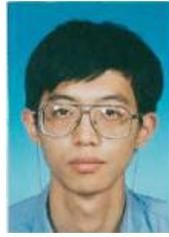

**Wei Hui** received the Ph.D. degree at the department of computer science at Beijing University of Aeronautics and Astronautics in 1998. From 1998 to 2000, he was a postdoctoral fellow at the department of computer science and the institute of artificial intelligence at Zhejiang University. Since November 2000, he has joined the department of computer science and engineering at Fudan University. His research interests include neuro-inspired artificial intelligence and cognitive science.

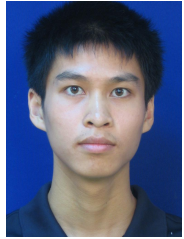

**Tang Fuyu**, born in 1994, Master of Computer Science, Fudan University. His research interest is image understanding.